\documentclass[runningheads]{llncs}
\usepackage[T1]{fontenc}
\usepackage{amsmath}
\usepackage{algorithm}
\usepackage{algpseudocode}
\usepackage{graphicx}
\usepackage{booktabs}
\usepackage{multirow}
\usepackage{makecell}
\usepackage[hidelinks]{hyperref}
\usepackage{xurl}

\begin{document}
\title{VCDF: A Validated Consensus-Driven Framework for Time Series Causal Discovery}
\titlerunning{VCDF for Time Series Causal Discovery}

\author{Gene Yu\inst{1}\orcidID{0000-0003-4867-1340}\thanks{Corresponding author} \and
Ce Guo\inst{1}\orcidID{0000-0002-0272-9175} \and
Wayne Luk\inst{1}\orcidID{0000-0002-6750-927X}}
\authorrunning{G. Yu et al.}

\institute{
Department of Computing, Imperial College London, London, United Kingdom \\
\email{gene.yu23@alumni.imperial.ac.uk, c.guo@imperial.ac.uk, w.luk@imperial.ac.uk}
}

% --- Double-blind anonymous setup ---
% \author{Anonymous Author(s)}
% \authorrunning{Anonymous}
% \institute{Paper under double-blind review}

%
\maketitle              % typeset the header of the contribution
\begin{abstract}
Time series causal discovery is essential for understanding dynamic systems, yet many existing methods remain sensitive to noise, non-stationarity, and sampling variability. We propose the Validated Consensus-Driven Framework (VCDF), a simple and method-agnostic layer that improves robustness by evaluating the stability of causal relations across blocked temporal subsets. VCDF requires no modification to base algorithms and can be applied to methods such as VAR-LiNGAM and PCMCI.
Experiments on synthetic datasets show that VCDF improves VAR-LiNGAM by approximately 0.08–0.12 in both window and summary F1 scores across diverse data characteristics, with gains most pronounced for moderate-to-long sequences. The framework also benefits from longer sequences, yielding up to 0.18 absolute improvement on time series of length 1000 and above. Evaluations on simulated fMRI data and IT-monitoring scenarios further demonstrate enhanced stability and structural accuracy under realistic noise conditions.
VCDF provides an effective reliability layer for time series causal discovery without altering underlying modeling assumptions.

\end{abstract}
\section{Introduction}

Causal discovery for multivariate time series plays a central role in understanding dynamic systems across domains such as finance \cite{Granger1969,Kristjanpoller2025}, neuroscience \cite{Smith2011}, and climate science \cite{Runge2023}. By identifying directional relations and temporal dependencies, causal models support downstream analyses including forecasting, diagnosis, and system interpretation. However, real-world time series frequently exhibit non-stationarity, non-linearity, heterogeneous noise, and distributional shifts \cite{Huang2020}. In addition, many systems involve large numbers of interacting components whose complex network structure further complicates causal inference \cite{Boccaletti2006}. These characteristics make reliable structural estimation difficult.

Representative causal discovery algorithms such as VAR-LiNGAM and PCMCI remain widely used due to their interpretability and theoretical grounding \cite{Hyvarinen2010,Runge2019}. Yet extensive evaluations indicate that their outputs may vary when noise levels, temporal regimes, or other data characteristics change \cite{Runge2019,Assaad2022}. Sensitivity to sampling variability and structural assumptions can lead to inconsistent causal graphs across data subsets \cite{Assaad2022,Glymour2019}. This instability undermines reliability and challenges practical adoption in applications that require consistent structural estimates.

A key observation motivating this work is that such variability across data partitions is itself informative. Relations that consistently appear across multiple subsets are more likely to reflect genuine causal structure, whereas highly variable edges may arise from estimation noise or structural ambiguity. Similar ideas appear in robustness studies and ensemble-based approaches, where agreement across models is used as a measure of reliability \cite{Chen2009}. However, the implications of these principles for validating causal structures in time series remain under-explored.

We introduce the \textbf{Validated Consensus-Driven Framework (VCDF)}, a simple and method-agnostic validation layer that evaluates the stability of causal relations across blocked temporal partitions. VCDF computes directional consistency and variability metrics for each candidate edge and removes those that fail to meet stability thresholds. Because VCDF only requires the base algorithm to output directed relations with an associated score or effect estimate, it can be applied to diverse causal discovery paradigms—including linear non-Gaussian models, constraint-based methods, and non-linear approaches—without modifying their modeling assumptions.

Extensive experiments on synthetic data demonstrate that VCDF improves structural accuracy across a wide range of temporal dynamics, noise conditions, and data-generating processes. Benefits are particularly pronounced for long sequences, where consensus validation can exploit richer temporal evidence. Experiments on simulated fMRI data and IT-monitoring scenarios further show that VCDF enhances robustness in complex real-world settings.

Our main contributions are:
\begin{itemize}
    \item \textbf{Consensus-driven refinement:} A validation framework that assesses the stability of causal relations across blocked data partitions.
    \item \textbf{Method-agnostic integration:} VCDF applies to a broad range of causal discovery algorithms without altering their assumptions.
    \item \textbf{Comprehensive empirical evaluation:} Across synthetic, fMRI, and IT-monitoring datasets, VCDF consistently improves both window and summary F1 scores.
\end{itemize}

\section{Background and Related Work}

\subsection{Time Series Causal Discovery}

Causal discovery in multivariate time series aims to uncover lagged and contemporaneous dependencies that govern system behavior \cite{Yao2021}. Two complementary representations are commonly used: \emph{window causal graphs}, which model lag-specific effects, and \emph{summary causal graphs}, which aggregate influences across all lags \cite{Assaad2022}. Compared with static causal inference, the temporal dimension introduces additional challenges, including variable delays, evolving relationships, and indirect pathways \cite{Runge2019,Spirtes2001}. These challenges are amplified in the presence of non-stationarity, non-linear relations, non-Gaussian disturbances, and heterogeneous temporal regimes \cite{Huang2020,Zhang2017}. High dimensionality further increases estimation noise and computational complexity \cite{Colombo2012}.

\subsection{Representative Methods}

A range of methods has been developed for time-series causal discovery. VAR-LiNGAM extends linear non-Gaussian acyclic models with vector autoregression, enabling identification of linear causal structures under non-Gaussian noise \cite{Hyvarinen2010}. PCMCI and related approaches use conditional independence tests and momentary conditional independence measures to handle high-dimensional and partially non-linear dependencies \cite{Runge2019,Runge2020}. Neural and optimization-based methods such as TCDF and DYNOTEARS extend causal modeling capabilities to non-linear architectures and continuous optimization formulations \cite{Nauta2019,Pamfil2020}. Despite their differences in modeling assumptions, these methods remain sensitive to noise, temporal heterogeneity, and sampling variability \cite{Mooij2016,Glymour2019}.

\subsection{Robustness and Stability}

Achieving robustness in causal structure estimation remains a significant challenge. Small perturbations in data, distributional shifts, or changes in sampling can lead to substantial variations in inferred graphs \cite{Assaad2022}. This instability is particularly problematic in time series, where dependencies span multiple lags and temporal regimes \cite{Zhang2017}. Recent studies highlight the need for scalable and reliable methods for practical deployment \cite{Hagedorn2022,Hagedorn2023,Guo2022,Guo2023}. The idea of leveraging consistency across models also appears in ensemble and robustness research, such as negative-correlation learning \cite{Chen2009}, where agreement across multiple predictors is used as a measure of reliability. These insights motivate validation strategies that assess stability across temporal partitions.

\subsection{Cross-Validation for Time Series}

Cross-validation is widely used to evaluate generalization in predictive modeling \cite{Stone1974,Kohavi1995}. However, time-series data require preserving temporal order, as random shuffling breaks autocorrelation structures \cite{Arlot2010}. Blocked K-fold cross-validation has been shown to be appropriate for autoregressive processes under mild assumptions \cite{Bergmeir2018}. While existing approaches focus primarily on prediction error, causal discovery requires validating the \emph{structural} stability of inferred relations. This motivates VCDF, which uses blocked folds not for model selection or prediction assessment, but for edge-stability validation. In each run, one contiguous block is held out and the base method is applied to the remaining data, yielding leave-one-block-out causal estimates. VCDF retains relations that are stable across runs, measured by $C(r_{ij})$ and $V(r_{ij})$.

\section{Validated Consensus-Driven Framework}

\subsection{Motivation}

\begin{figure}[t]
    \centering
    \includegraphics[width=\textwidth]{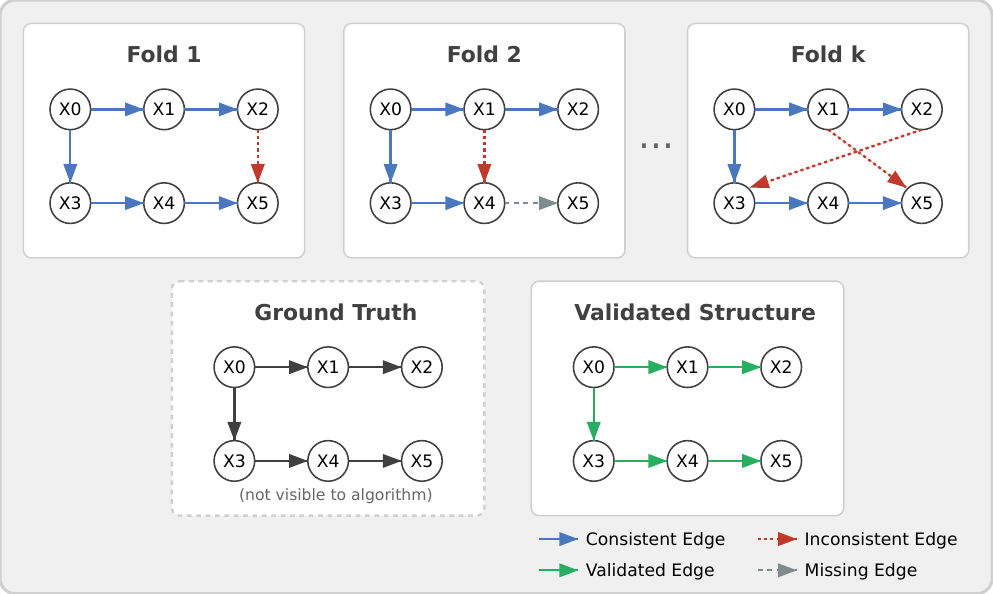}
    \caption{Illustration of fold-to-fold variability and validated edges in VCDF.
    Consistent relations across partitions are retained; unstable ones are removed.}
    \label{fig:vcdf_validation}
\end{figure}

Outputs of causal discovery methods can vary substantially when data are noisy, non-stationary, or sampled differently over time. Applying the same algorithm to different partitions of the same dataset often yields inconsistent edges due to sensitivity to sampling variation. We treat this variability as informative: relations that appear consistently across subsamples are more likely to reflect genuine causal structure, whereas unstable edges may reflect estimation noise or structural ambiguity. This motivates a validation procedure that examines how each relation behaves across multiple blocked partitions of the data, as illustrated in Fig.~\ref{fig:vcdf_validation}.

\begin{figure}[htb]
    \centering
    \includegraphics[width=0.85\textwidth]{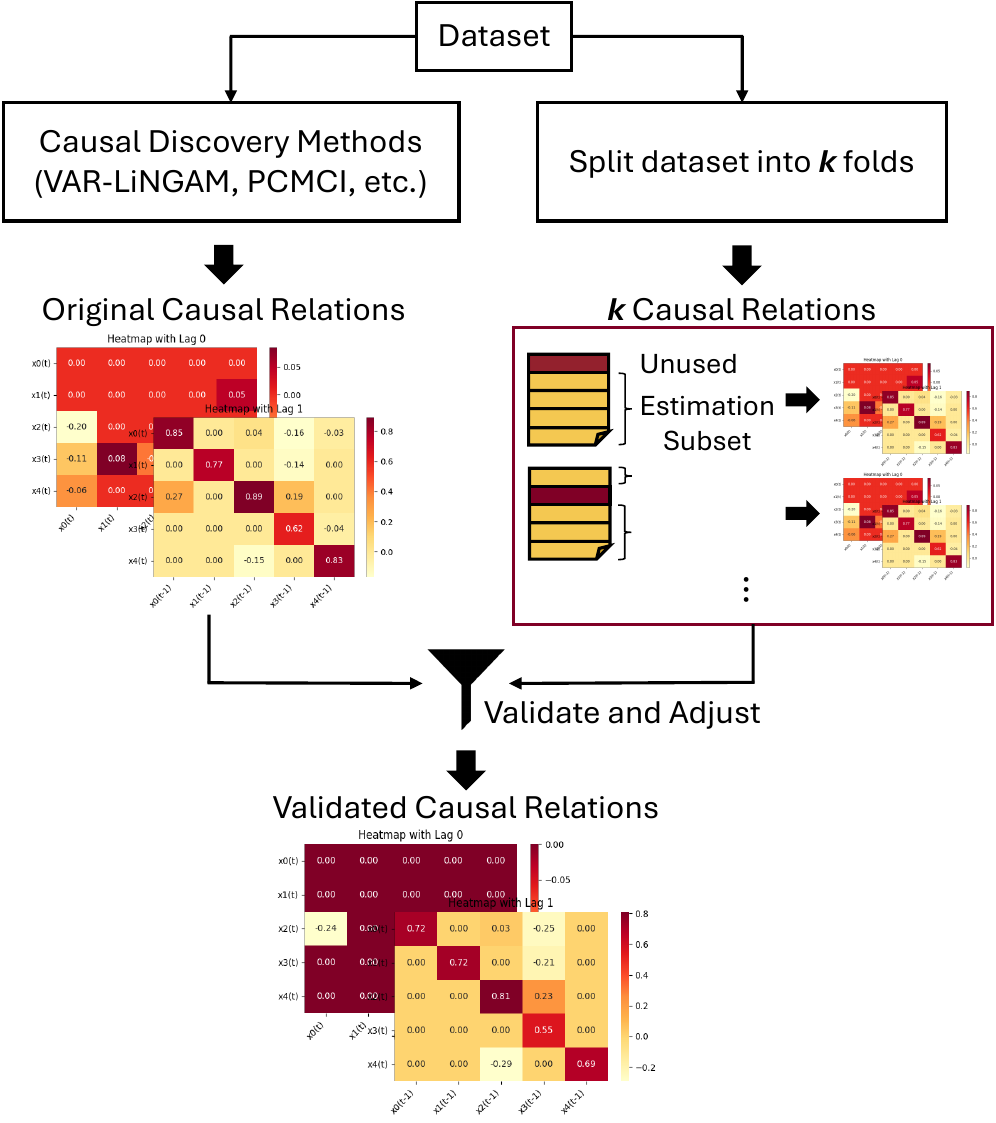}
    \caption{Workflow of VCDF. Each fold produces a causal estimate; stability
    validation aggregates them into a final graph.}
    \label{fig:vcdf_workflow}
\end{figure}

\subsection{Blocked Partitioning}
The overall workflow of VCDF is shown in Fig.~\ref{fig:vcdf_workflow}. 
Given a multivariate time series $\mathcal{D}$, VCDF divides it into $k$ contiguous folds. For fold $m$, the held-out block $\mathcal{D}_m^{\mathrm{val}}$ is excluded, and the remaining segments form the training set:
\[
\mathcal{D}_m^{\mathrm{train}} = \mathcal{D} \setminus \mathcal{D}_m^{\mathrm{val}}.
\]
Temporal order is preserved, and the validation block is never used during estimation. Running a base causal discovery method (e.g., VAR-LiNGAM or PCMCI) independently on each $\mathcal{D}_m^{\mathrm{train}}$ yields graphs $\{G_1,\dots,G_k\}$ as well as the full-sample estimate $G_0$. Although removing a contiguous block may create a small number of cross-boundary pairs when concatenating segments, their impact is negligible when the maximum lag is small relative to sequence length.

\subsection{Stability Metrics}

Let $r_{ij}^0$ denote the effect estimate for edge $i \rightarrow j$ from the full dataset, and $r_{ij}^m$ the corresponding estimate from fold $m$. VCDF evaluates each relation with two metrics. First, directional consistency:
\[
C(r_{ij}) = \frac{1}{k} \sum_{m=1}^k \mathbf{1}\{\operatorname{sign}(r_{ij}^m) = \operatorname{sign}(r_{ij}^0)\},
\]
and second, relative variability:
\[
V(r_{ij}) = \frac{\operatorname{std}(\{r_{ij}^m\})}{|r_{ij}^0| + \epsilon}.
\]
VCDF assumes that effect estimates are comparable across folds. Stable relations should exhibit high directional consistency and low normalized variability.

\subsection{Filtering and Optional Adjustment}

VCDF retains only relations that satisfy the stability thresholds:
\[
C(r_{ij}) \ge \tau_c, \qquad V(r_{ij}) \le \tau_v.
\]
In our experiments, thresholds are set as method-specific fixed defaults selected once via a synthetic sweep and then kept unchanged across datasets:
for VARLiNGAM, $\tau_c{=}0.4,\tau_v{=}0.4$; for PCMCI, $\tau_c{=}0.7,\tau_v{=}0.4$. Optionally, effect-size refinement interpolates between the full-sample estimate and the cross-fold mean:
\[
r_{ij}^{\mathrm{adj}}
    = (1-w)\, r_{ij}^0 + w \cdot \operatorname{mean}(\{r_{ij}^m\}),
\]
where $w \in [0,1]$. In this work we use $w=0$ to focus on structural validation; effect-size refinement is optional when magnitude estimation is of interest.

\subsection{Algorithm}

The procedure is summarized in Algorithm~1. VCDF produces a refined version of $G_0$ by removing unstable edges but does not introduce new ones, as it relies on signals supported by the full dataset.

\begin{algorithm}[tb]
\caption{Validated Consensus-Driven Framework (VCDF)}
\begin{algorithmic}[1]
\Require Dataset $\mathcal{D}$, folds $k$, thresholds $\tau_c,\tau_v$
\State $G_0 \gets$ BaseCausalDiscovery($\mathcal{D}$)
\For{$m=1$ to $k$}
    \State Split $\mathcal{D}$ into $\mathcal{D}_m^{\mathrm{train}}$ and $\mathcal{D}_m^{\mathrm{val}}$
    \State $G_m \gets$ BaseCausalDiscovery($\mathcal{D}_m^{\mathrm{train}}$)
\EndFor
\For{each relation $i \rightarrow j$ in $G_0$}
    \State Compute $C(r_{ij})$ and $V(r_{ij})$
    \If{$C(r_{ij}) < \tau_c$ or $V(r_{ij}) > \tau_v$}
        \State Remove $i \rightarrow j$ from $G_0$
    \EndIf
\EndFor
\State \Return $G_0$
\end{algorithmic}
\end{algorithm}

VCDF introduces moderate runtime overhead that scales approximately linearly with $k$ and depends primarily on the cost of the base causal discovery method. In typical settings, $k$ ranges from 5 to 10, offering a practical balance between stability and computational cost.

\section{Experimental Evaluation}

We evaluate VCDF on both synthetic and domain-specific time-series datasets. The goal is to assess whether consensus-based validation improves structural accuracy across diverse data characteristics and temporal scales. Following Assaad et al.~\cite{Assaad2022}, we report both window F1 (lag-specific) and summary F1 (time-aggregated) scores as standard structural metrics, while additional measures show consistent trends.

\begin{figure}[tb]
    \centering
    \includegraphics[width=0.95\linewidth]{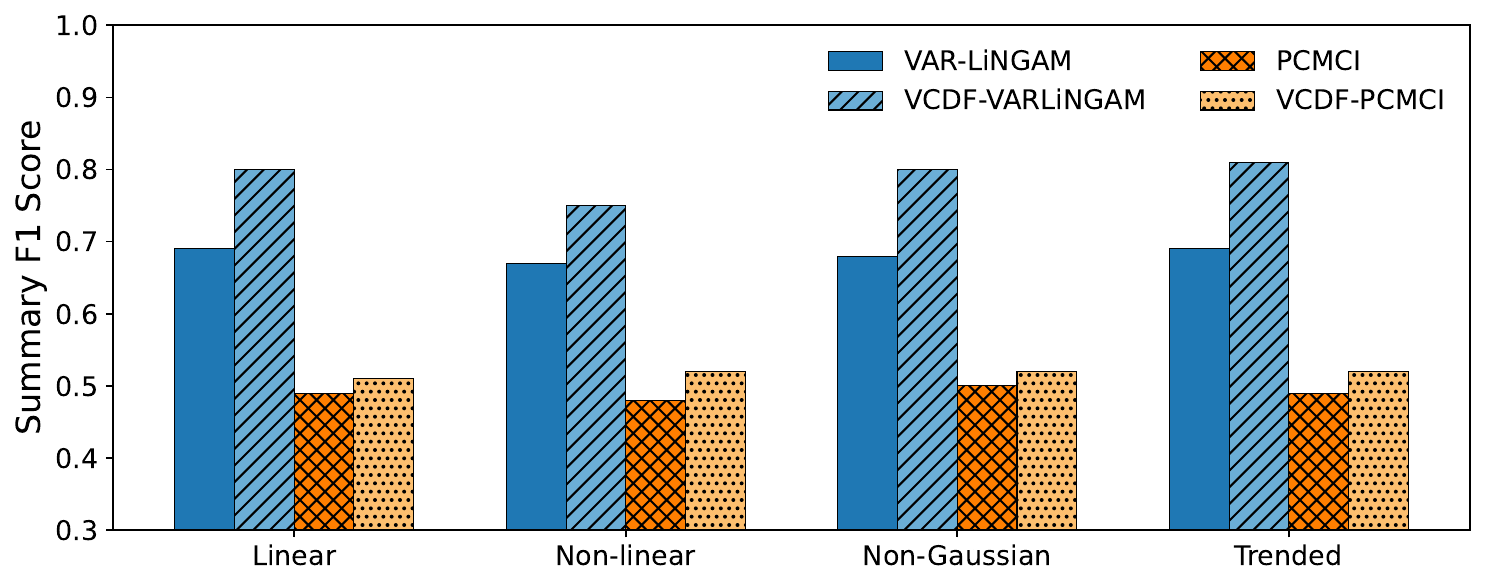}
    \caption{
    Comparison of VAR-LiNGAM, VCDF-VARLiNGAM, PCMCI, and VCDF-PCMCI across representative synthetic settings.
    }
    \label{fig:vcdf_comparison}
\end{figure}

\subsection{Synthetic Data}

We use four representative synthetic settings widely adopted in causal discovery benchmarks, capturing (1) linear vs.\ nonlinear relations, (2) Gaussian vs.\ non-Gaussian errors, and (3) trended vs.\ stationary temporal dynamics. Each configuration contains 15 variables with multiple dataset realizations.

We compare VCDF against four causal discovery paradigms: VAR-LiNGAM, PCMCI, TCDF, and DYNOTEARS. Since TCDF and DYNOTEARS rely on regularization or neural filtering, their variability across folds is smaller; thus, we focus detailed analysis on VAR-LiNGAM and PCMCI, where consensus validation yields clearer improvements.

Table~\ref{tab:characteristics} summarises results across characteristics. VCDF consistently boosts both window and summary F1 scores. In the linear setting, VCDF-VARLiNGAM improves summary F1 from $0.69$ to $0.80$. Across all four characteristics, VARLiNGAM gains $0.08$–$0.12$ summary F1 after applying VCDF. PCMCI also benefits from consensus filtering, with summary F1 improvements of $0.02$--$0.04$ across all characteristics. These results confirm that VCDF effectively filters unstable edges while preserving true causal relations.

\begin{table}[tb]
  \caption{Performance across four representative synthetic settings.}
  \label{tab:characteristics}
  \centering
  \small
  \setlength{\tabcolsep}{4pt}
  \begin{tabular}{c@{\hspace{4pt}}l@{\hspace{12pt}}cc}
    \toprule
    \multirow{2}{*}{Setting} & \multirow{2}{*}{Method} & \multicolumn{2}{c}{F1 score} \\
    \cmidrule(lr){3-4}
    & & Window & Summary \\
    \midrule

    % -------- Linear (S1) ----------
    \multirow{6}{*}{Linear} 
      & PCMCI           & 0.37 $\pm$ 0.02 & 0.49 $\pm$ 0.03 \\
      & VCDF-PCMCI      & 0.43 $\pm$ 0.02 & 0.51 $\pm$ 0.02 \\
      & VARLiNGAM       & 0.64 $\pm$ 0.03 & 0.69 $\pm$ 0.03 \\
      & \textbf{VCDF-VARLiNGAM} & \textbf{0.75 $\pm$ 0.03} & \textbf{0.80 $\pm$ 0.03} \\
      & TCDF            & 0.38 $\pm$ 0.03 & 0.43 $\pm$ 0.03 \\
      & DYNOTEARS       & 0.48 $\pm$ 0.04 & 0.53 $\pm$ 0.04 \\
    \midrule

    % -------- Non-linear (S2) ----------
    \multirow{6}{*}{Non-linear} 
      & PCMCI           & 0.36 $\pm$ 0.01 & 0.48 $\pm$ 0.01 \\
      & VCDF-PCMCI      & 0.44 $\pm$ 0.02 & 0.52 $\pm$ 0.02 \\
      & VARLiNGAM       & 0.61 $\pm$ 0.04 & 0.67 $\pm$ 0.04 \\
      & \textbf{VCDF-VARLiNGAM} & \textbf{0.69 $\pm$ 0.03} & \textbf{0.75 $\pm$ 0.03} \\
      & TCDF            & 0.35 $\pm$ 0.01 & 0.38 $\pm$ 0.01 \\
      & DYNOTEARS       & 0.54 $\pm$ 0.04 & 0.60 $\pm$ 0.03 \\
    \midrule

    % -------- Non-Gaussian (S4) ----------
    \multirow{6}{*}{Non-Gaussian} 
      & PCMCI           & 0.38 $\pm$ 0.02 & 0.50 $\pm$ 0.02 \\
      & VCDF-PCMCI      & 0.43 $\pm$ 0.02 & 0.52 $\pm$ 0.03 \\
      & VARLiNGAM       & 0.62 $\pm$ 0.04 & 0.68 $\pm$ 0.05 \\
      & \textbf{VCDF-VARLiNGAM} & \textbf{0.74 $\pm$ 0.04} & \textbf{0.80 $\pm$ 0.04} \\
      & TCDF            & 0.40 $\pm$ 0.04 & 0.45 $\pm$ 0.04 \\
      & DYNOTEARS       & 0.47 $\pm$ 0.04 & 0.52 $\pm$ 0.04 \\
    \midrule

    % -------- Trended (S6) ----------
    \multirow{6}{*}{Trended} 
      & PCMCI           & 0.38 $\pm$ 0.02 & 0.49 $\pm$ 0.02 \\
      & VCDF-PCMCI      & 0.42 $\pm$ 0.03 & 0.52 $\pm$ 0.02 \\
      & VARLiNGAM       & 0.64 $\pm$ 0.03 & 0.69 $\pm$ 0.03 \\
      & \textbf{VCDF-VARLiNGAM} & \textbf{0.76 $\pm$ 0.04} & \textbf{0.81 $\pm$ 0.04} \\
      & TCDF            & 0.36 $\pm$ 0.03 & 0.40 $\pm$ 0.04 \\
      & DYNOTEARS       & 0.46 $\pm$ 0.05 & 0.51 $\pm$ 0.05 \\
    \bottomrule
  \end{tabular}
\end{table}

Figure~\ref{fig:vcdf_comparison} visualizes these improvements. VCDF-VARLiNGAM shows the largest gains across settings, while VCDF-PCMCI reliably enhances its baseline.

\subsubsection{Effect of Sequence Length.}
To evaluate temporal scalability, we vary sequence length from $T=250$ to $T=2000$. Table~\ref{tab:temporal} shows that baseline VAR-LiNGAM saturates beyond $T \approx 500$, whereas VCDF-VARLiNGAM continues to improve, reaching window F1 $\approx 0.79$ and summary F1 $\approx 0.82$--$0.83$ at $T=1000$--$2000$. These results indicate that VCDF leverages additional temporal information more effectively, enhancing robustness without overfitting to noise. For very short sequences (e.g., $T{=}250$), base methods may be underpowered, and VCDF's estimates are computed from even shorter effective samples, which can reduce the benefit of stability filtering and occasionally lead to slight degradation.

\begin{table}[tb]
\caption{Performance across different time-series lengths (synthetic, $n=15$).}
\label{tab:temporal}
\centering
\small
\begin{tabular}{c@{\hspace{8pt}}l@{\hspace{12pt}}cc}
\toprule
\multirow{2}{*}{Length $T$} & \multirow{2}{*}{Method} & \multicolumn{2}{c}{F1 score} \\
\cmidrule(lr){3-4}
& & Window & Summary \\
\midrule
\multirow{3}{*}{250} 
 & VARLiNGAM        & 0.65 $\pm$ 0.05 & \textbf{0.71 $\pm$ 0.03} \\
 & VCDF-VARLiNGAM   & \textbf{0.65 $\pm$ 0.04} & 0.67 $\pm$ 0.03 \\
 & PCMCI            & 0.39 $\pm$ 0.03 & 0.52 $\pm$ 0.03 \\
\midrule
\multirow{3}{*}{1000}
 & VARLiNGAM        & 0.64 $\pm$ 0.02 & 0.71 $\pm$ 0.03 \\
 & \textbf{VCDF-VARLiNGAM}   & \textbf{0.79 $\pm$ 0.04} & \textbf{0.82 $\pm$ 0.04} \\
 & PCMCI            & 0.38 $\pm$ 0.02 & 0.53 $\pm$ 0.01 \\
\midrule
\multirow{3}{*}{2000}
 & VARLiNGAM        & 0.61 $\pm$ 0.03 & 0.66 $\pm$ 0.02 \\
 & \textbf{VCDF-VARLiNGAM}   & \textbf{0.79 $\pm$ 0.02} & \textbf{0.83 $\pm$ 0.02} \\
 & PCMCI            & 0.37 $\pm$ 0.03 & 0.52 $\pm$ 0.03 \\
\bottomrule
\end{tabular}
\end{table}

\subsection{Domain-Specific Datasets}

We further evaluate VCDF on simulated fMRI data~\cite{Smith2011} and three IT monitoring systems (antivirus, middleware, and web services). Ground-truth causal structures correspond to system topology, enabling reliable evaluation.

On the fMRI benchmark, VCDF-PCMCI achieves the best summary F1, while VCDF-VARLiNGAM improves temporal precision with the highest window F1. In IT-monitoring datasets, VCDF-VARLiNGAM improves summary F1 by approximately 10--15\% across all systems, demonstrating that consensus-based refinement remains effective under realistic, noisy operational conditions.

\begin{table}[tb]
\caption{Performance on domain-specific datasets (summary F1 shown where applicable).}
\label{tab:real_datasets}
\centering
\small
\begin{tabular}{c@{\hspace{8pt}}l@{\hspace{12pt}}cc}
\toprule
\multirow{2}{*}{Dataset} & \multirow{2}{*}{Method} & \multicolumn{2}{c}{F1 score} \\
\cmidrule(lr){3-4}
& & Window & Summary \\
\midrule
\multirow{3}{*}{fMRI}
 & VARLiNGAM         & 0.48 $\pm$ 0.07 & 0.61 $\pm$ 0.12 \\
 & VCDF-VARLiNGAM    & \textbf{0.51 $\pm$ 0.14} & 0.57 $\pm$ 0.17 \\
 & VCDF-PCMCI        & 0.39 $\pm$ 0.12 & \textbf{0.64 $\pm$ 0.11} \\
\midrule
\multirow{3}{*}{IT Antivirus}
 & VARLiNGAM         & \multicolumn{1}{c}{---} & 0.35 \\
 & VCDF-VARLiNGAM    & \multicolumn{1}{c}{---} & \textbf{0.49} \\
 & VCDF-PCMCI        & \multicolumn{1}{c}{---} & 0.36 \\
\midrule
\multirow{3}{*}{IT MoM}
 & VARLiNGAM         & \multicolumn{1}{c}{---} & 0.44 \\
 & VCDF-VARLiNGAM    & \multicolumn{1}{c}{---} & \textbf{0.56} \\
 & VCDF-PCMCI        & \multicolumn{1}{c}{---} & 0.47 \\
\midrule
\multirow{3}{*}{IT Web}
 & VARLiNGAM         & \multicolumn{1}{c}{---} & 0.45 \\
 & VCDF-VARLiNGAM    & \multicolumn{1}{c}{---} & \textbf{0.55} \\
 & VCDF-PCMCI        & \multicolumn{1}{c}{---} & 0.47 \\
\bottomrule
\end{tabular}
\end{table}

\subsection{Runtime Considerations}

VCDF incurs moderate computational overhead due to repeated estimation on $k$ partitions. Runtime grows approximately linearly with $T$ and $k$. For 15-variable systems with $T=1000$, VCDF-VARLiNGAM runs in several seconds, remaining practical for typical time-series analysis. Table~\ref{tab:runtime} reports runtime comparisons.

\begin{table}[t]
\caption{Runtime comparison (seconds) on synthetic datasets with 15 variables.}
\label{tab:runtime}
\centering
\small
\begin{tabular}{ccc}
\toprule
Length $T$ & VAR-LiNGAM & VCDF-VARLiNGAM \\
\midrule
250  & 0.50 & 2.83 \\
500  & 0.65 & 3.62 \\
1000 & 0.96 & 5.15 \\
2000 & 1.75 & 8.66 \\
\bottomrule
\end{tabular}
\end{table}

\subsection{Summary}

Across synthetic and domain-specific datasets, VCDF consistently improves the stability and accuracy of time-series causal discovery. Synthetic benchmarks show that, for VAR-LiNGAM, VCDF yields gains of 0.08–0.12 in both window and summary F1 scores across linear, nonlinear, non-Gaussian, and trended settings. VCDF benefits particularly from longer sequences, where baseline methods tend to saturate. Results on fMRI and IT-monitoring datasets further confirm that consensus filtering improves robustness under realistic noise conditions. Overall, VCDF acts as an effective reliability layer that complements a range of causal discovery algorithms.

\section{Conclusion}

We introduced VCDF, a consensus-driven validation framework that enhances the robustness of time-series causal discovery by evaluating stability across blocked data partitions. As a model-agnostic post-processing layer, VCDF can be applied to existing algorithms without altering their assumptions, and improves both window and summary F1 scores across diverse synthetic conditions, particularly for moderate-to-long sequences. Experiments on fMRI and IT-monitoring datasets further demonstrate its practical effectiveness in noisy, real-world-like settings. VCDF is a validation layer, so recall is bounded by the base method. When causal structure genuinely changes over time, cross-partition inconsistency may reflect true change rather than noise.

These results suggest that consensus validation provides a simple yet reliable mechanism for strengthening causal discovery pipelines. Future work includes extending VCDF to scenarios with latent confounders, exploring adaptive stability thresholds, and developing more scalable implementations for high-dimensional or streaming data.

Code is available at \url{https://github.com/sonnets-project/vcdf-time-series-causal-discovery}.

\begin{credits}
\subsubsection{\ackname} We thank the anonymous reviewers and the dataset providers. We are also grateful to Dr Paul Labonne (Bank of England) for his valuable feedback. This work is supported by the United Kingdom EPSRC (grant numbers UKRI256, EP/V028251/1, EP/N031768/1, EP/S030069/1, and EP/X036006/1), Altera, Intel, AMD, and SiliconFlow.

% \subsubsection{\discintname}
% The authors have no competing interests to declare that are
% relevant to the content of this article.
\end{credits}

\bibliographystyle{splncs04}
\bibliography{main-refs}
\end{document}